\documentclass[letterpaper, 10 pt, conference]{ieeeconf}  

\IEEEoverridecommandlockouts             
\usepackage{graphicx}
\usepackage{url}
\usepackage{slashbox}
\usepackage{amsmath, amssymb}
\usepackage{algorithmic}
\usepackage{algorithm}

\title{\LARGE \bf Applying Rule-Based Context Knowledge to Build Abstract Semantic Maps of Indoor Environments}

\author{\parbox{5 in}{\centering Ziyuan Liu\quad\quad\quad\quad Georg von Wichert
\thanks{Z. Liu is with the Institute of Automatic Control Engineering, Technische Universit\"at M\"unchen, D-80290, Munich, Germany. \small{\texttt{ziyuan.liu@tum.de}}}
\thanks{G. von Wichert is with Siemens AG, Corporate Technology, Munich, Germany and Institute for Advanced Study, Techniche Universit\"at M\"unchen, Munich, Germany \small{\texttt{georg.wichert@siemens.com}}  }
}
}

\begin{document}

\maketitle
\thispagestyle{empty}
\pagestyle{empty}

\begin{abstract}
In this paper, we propose a generalizable method that systematically combines data driven MCMC sampling and inference using rule-based context knowledge for data abstraction. In particular, we demonstrate the usefulness of our method in the scenario of building abstract semantic maps for indoor environments. The product of our system is a parametric abstract model of the perceived environment that not only accurately represents the geometry of the environment but also provides valuable abstract information which benefits high-level robotic applications. Based on predefined abstract terms, such as ``type" and ``relation", we define task-specific context knowledge as descriptive rules in Markov Logic Networks. The corresponding inference results are used to construct a prior distribution that aims to add reasonable constraints to the solution space of semantic maps. In addition, by applying a semantically annotated sensor model, we explicitly use context information to interpret the sensor data. Experiments on real world data show promising results and thus confirm the usefulness of our system.
\end{abstract}

\section{Introduction}
\label{int}

In recent years, the performance of autonomous systems has been greatly improved. Multicore CPUs, bigger RAMs, new sensors and faster data flow have made many applications possible which seemed to be unrealistic in the past. However, the performance of such systems tends to become quite limited, as soon as they leave their carefully engineered operating environments. On the other hand, people may ask, why we humans can handle highly complex problems. Maybe the exact answer to this question still remains unclear, however, it is obvious that abstraction and knowledge together play an important role. We humans understand the world in abstract terms and have the necessary knowledge, based on which we can make inference given only partial information. As a human, if we see a desk in an office room, instead of memorizing the world coordinates of all the surface points of the desk, we will only notice that there is an object ``desk" at a certain position, and even this position is probably described in abstract terms like ``beside the window" or ``near the door". Based on prior knowledge, we can make some reasonable assumptions, such as there could be some ``books" in the ``drawer" of the desk, instead of some ``shoes" being inside, without opening the drawer. In our work, we aim to deploy such abilities (abstraction and inference) in the area of semantic robot mapping.


\section{Related Work}
In general, related work on semantic robot mapping can be classified into several groups. A big body of literature focuses on semantic place labelling which divides the environment into several regions and assigns each region a semantic label, such as ``office room" or ``corridor". Park and Song \cite{park2011hybrid} proposed a hybrid semantic mapping system for home environments, explicitly using information about doors as a key feature. Combining image segmentation and object recognition, Jebari et. al. \cite{jebari2011multi} extended semantic place labelling with object detection. Based on human augmented mapping, rooms and hallways are represented as Gaussian distributions to help robot navigate in \cite{Nieto-Granda2010}. Pronobis and Jensfelt \cite{pronobis2012large} integrated multi-modal sensory information and human intervention to classify places with semantic types. Other examples on semantic place labelling can be found in \cite{goerke2009building}, \cite{krishnan2010visual} and \cite{sjoo2012semantic}.


Different from place labelling,  another group of work concentrates on labelling different parts of the perceived environments with semantic tags, such as walls, floors, ceilings of indoor environments, or buildings, roads, vegetations of outdoor environments. In \cite{nuechter2008towards}, a logic-based constraint network describing the relations between different parts is used for labelling indoor environments. Persson and Duckett \cite{persson2007probabilistic} combined range data and omni-directional images to detect outlines of buildings and natural objects in an outdoor setting. Other examples in this category can be found in \cite{shim20113d}, \cite{an2012} and \cite{wolf2008semantic}.


\indent Another category consists of object-based semantic mapping systems which use object as basic representation unit of the perceived environment. Such systems usually adopt point cloud processing and image processing techniques to model or detect objects. Object features like appearance, shape and 3D locations are used to represent the objects. Examples of object-based semantic mapping can be found in \cite{rusu2009model}, \cite{ranganathan2007semantic}, \cite{pangercicsemantic} and \cite{masonobject}.


\indent In this paper, we extend our previous work \cite{liu2013extracting} using rule-based context knowledge. The work as a whole demonstrates a probabilistic method for building abstract semantic maps for indoor environments, which systematically combines data driven MCMC \cite{tu2005image} and inference using rule-based context knowledge. Unlike semantic labelling processes, whose typical output is a map data set with semantic tags, our mapping system outputs a parametric abstract model of the perceived environment, which not only accurately represents the geometry of the environment but also provides valuable abstract information.



\section{An Abstract Model for Semantic Indoor Maps}
\begin{figure}
	\centering
	\includegraphics[width=.9\columnwidth]{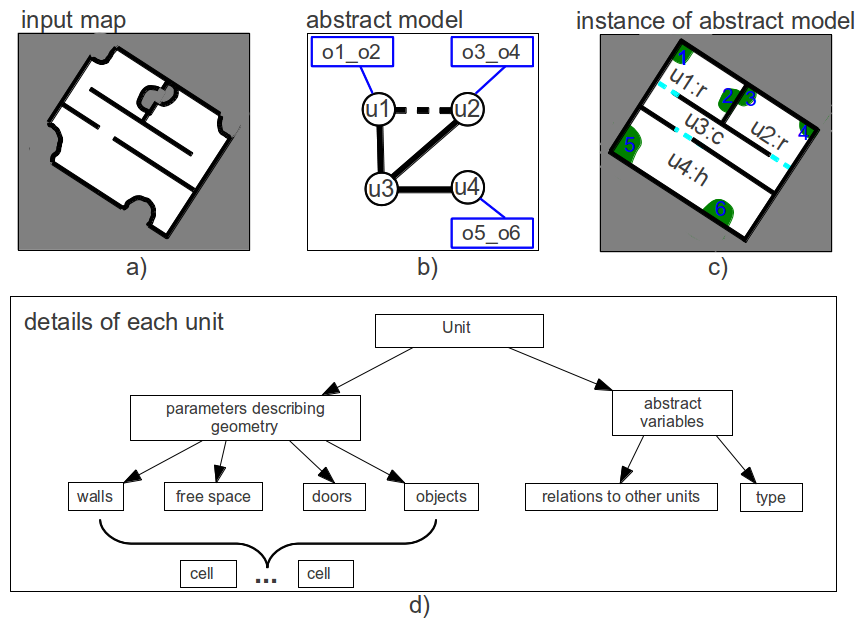}
    \caption{Overview of our abstract semantic map model. \textbf{a)} Input occupancy grid map (gray=unknown, white=free, black=occupied). \textbf{b)} The abstract model (circle nodes=space units, blue rectangle nodes=objects, black solid edges=connected by a door, black dashed edges=adjacent without door). \textbf{c)} An instance of the abstract model (gray=unknown, white=free, black=wall, cyan=door, green=object, r=room, c=corridor, h=hall) under the assumption that each space unit has a rectangular shape. \textbf{d)} Each unit in the abstract model contains abstract variables and parameters describing its geometry (size, position and orientation).}
    \label{figure:system overview}
\end{figure}

\indent Our semantic mapping system takes a grid map (a typical result of 2D SLAM processes, e.g. \cite{grisetti2007improved}) of the perceived environment as input and returns a parametric abstract model of this environment which provides semantic level explanation (such as ``type" and ``relation") and geometrical estimation of the environment. Explanation of our model is given in Fig. \ref{figure:system overview}. 

Our abstract model explains indoor environments in terms of basic indoor space types, such as ``room", ``corridor", ``hall" and so on, and we denote it as $W$:
\begin{equation}
W:=\{U,T,R\},
\label{eq:abstract model}
\end{equation}
where $U=\{u_i|i=1,\dots, n\}$ represents the set of all $n$ units. Each unit $u_i$ has a rectangle shape, and its geometry (size, position and orientation) is represented by its four vertices. The four edges of a unit are its walls. Doors are small line segments of free cells that are located in walls and connect to another unit. Unknown cells of the input map that are located within a unit are considered as object cells. All cells within a unit that do not belong to object cells are considered as free space of the unit. $T=\{t_i|i=1,\dots, n\}$ is the set of type of each individual unit, with $t_i\in\{\textrm{room,corridor,hall,other}\}$. Here ``other" indicates unit types that are not ``room", ``corridor" or ``hall". $R=\{r_{p,q}|p=1,\dots, n;q=1,\dots, n\}$ is a $n\times n$ matrix, whose element $r_{p,q}$ describes the relation between the unit $u_p$ and the unit $u_q$, with $r_{p,q}=r_{q,p}\in\{\neg \textrm{adjacent,adjacent}\}$. If two units share a wall, we define their relation as ``adjacent", otherwise ``$\neg$adjacent". By default, we define a unit $u_p$ is not adjacent to itself, i.e. $r_{p,p}=\neg\textrm{adjacent}$. In the following, we call each instance of the abstract model a ``semantic world" or ``world".

\indent A main criterion for evaluating how well a semantic world $W$ matches with the input grid map $M$ is the posterior probability $p(W|M)$, and it is computed as:
\begin{equation}
p(W|M)\propto p(M|W)\cdot p(W).
\label{eq:pos}
\end{equation}
Here, the term $p(M|W)$ is usually called \textit{likelihood} and indicates how probable the input is for different worlds. The term $p(W)$ is called \textit{prior} and describes the belief on which worlds are possible at all. In the following, we formulate task-specific context knowledge as descriptive rules in Markov Logic Networks (MLNs) \cite{richardson2006markov} and show how to define the likelihood and the prior using the inference results of MLNs in a systematic way. For details on MLNs, we refer to \cite{richardson2006markov}.

\subsection{Inference using rule-based context knowledge}
In general, context knowledge describes our prior belief for a certain domain, such as that the ground becomes wet after it has rained. Rather than exact quantitative information, context knowledge provides advisory qualitative information for our judgements. Such information is very valuable in handling problems of high dimensionality where computation suffers due to the huge state space. In the domain of robot indoor mapping, we formulate following context knowledge:
\begin{itemize}
\item There are four types of space units: room, corridor, hall and other.
\item Two units are either adjacent (neighbours) or not adjacent.
\item The type of a unit is dependent on its geometry and size.
\item In contrast to rooms, corridors have multiple doors.
\item Connecting walls of two adjacent rooms have the same length.
\end{itemize}
With the help of MLNs, we formulate our context knowledge as descriptive rules in Table \ref{TAB:formulas}. Based on these rules, query defined in Table \ref{TAB:query} can be made given evidence shown in Table \ref{TAB:evidence}. Using these rules, we try to formulate the features of certain indoor environments with rectangular space units. The choice of the rules is a problem-oriented engineering step, and the rules given in this paper serve as a good example.


\begin{table}[htb]
\centering
\begin{tabular}{|ll|}
\hline predicate&explanation \\ \hline
$RoLi(u_p)$&Unit $u_p$ has a room-like geometry.\\
${CoLi(u_p)}$&Unit $u_p$ has a corridor-like geometry.\\
${HaLi(u_p)}$&Unit $u_p$ has a hall-like geometry.\\
${MulDoor(u_p)}$&Unit $u_p$ has multiple doors.\\
${Adj(u_p,u_q)}$&Unit $u_p$ and $u_q$ are adjacent.\\
\hline 
\end{tabular} 
\caption{Definition of evidence predicates}
\label{TAB:evidence}     
\end{table}

\begin{table}[htb]
\centering
\begin{tabular}{|ll|}
\hline predicate&explanation \\ \hline
${Room(u_p)}$&Unit $u_p$ has the type of room.\\
${Corr(u_p)}$&Unit $u_p$ has the type of corridor.\\
${Hall(u_p)}$&Unit $u_p$ has the type of hall.\\
${Other(u_p)}$&Unit $u_p$ has the type of other.\\
${SaLe(u_p,u_q)}$&Unit $u_p$ and $u_q$ have each a \\&wall with the same length.\\
\hline 
\end{tabular} 
\caption{Definition of query predicates}
\label{TAB:query}     
\end{table}

\begin{table}[htb]
\centering
\begin{tabular}{|l|}
\hline basic features:\\
$Adj(u_p,u_q) \to Adj(u_q,u_p)$\\
$\textit{SaLe}(u_p,u_q) \to \textit{SaLe}(u_q,u_p)$\\
\hline
reasoning on type:\\
$ HaLi(u_p)\to Hall(u_p) $\\
$ HaLi(u_p)\to\neg Room(u_p) $\\
$ HaLi(u_p)\to\neg Corr(u_p) $\\
$ HaLi(u_p)\to\neg Other(u_p) $\\
$ RoLi(u_p)\to\neg Hall(u_p) $\\
$ CoLi(u_p)\to\neg Hall(u_p) $\\
$ RoLi(u_p)\land \neg MulDoor(u_p) \to Room(u_p) $\\
$ RoLi(u_p)\land \neg MulDoor(u_p) \to \neg Corr(u_p) $\\
$ RoLi(u_p)\land MulDoor(u_p) \to Other(u_p) $\\
$ CoLi(u_p)\land \neg MulDoor(u_p) \to Other(u_p)$\\
$ CoLi(u_p)\land MulDoor(u_p) \to Corr(u_p) $\\
$ CoLi(u_p)\land MulDoor(u_p) \to \neg Room(u_p) $\\\hline
reasoning on $SaLe$:\\
$\neg Adj(u_q,u_p)\to \neg \textit{SaLe}(u_p,u_q)$\\
$Room(u_p)\land Room(u_q) \land Adj(u_p,u_q)\to \textit{SaLe}(u_p,u_q) $\\
$Room(u_p)\land Hall(u_q) \land Adj(u_p,u_q)\to \neg \textit{SaLe}(u_p,u_q) $\\
$Room(u_p)\land Corr(u_q) \land Adj(u_p,u_q)\to \neg \textit{SaLe}(u_p,u_q) $\\
$Hall(u_p)\land Corr(u_q) \land Adj(u_p,u_q)\to \neg \textit{SaLe}(u_p,u_q) $\\
$Other(u_p)\land Hall(u_q) \land Adj(u_p,u_q)\to \neg \textit{SaLe}(u_p,u_q) $\\
$Other(u_p)\land Corr(u_q) \land Adj(u_p,u_q)\to \neg \textit{SaLe}(u_p,u_q) $\\
$Other(u_p)\land Room(u_q) \land Adj(u_p,u_q)\to \neg \textit{SaLe}(u_p,u_q) $\\
\hline 
\end{tabular} 
\caption{Context knowledge defined as descriptive rules}
\label{TAB:formulas}     
\end{table}

\indent Before we can make inference in MLNs, the evidence defined in Table \ref{TAB:evidence} need be provided as input for MLNs, which includes geometry evidence, relation evidence and evidence on doors. To provide the first, we use a classifier that categorizes the geometry of a unit into ``room-like", ``corridor-like" or ``hall-like" according to its size and length/width ratio. The general idea of this classifier is shown in Table \ref{TAB:classifier}.

\begin{table}[htb]
	\centering
	\begin{tabular}{|c||*{3}{c|}}\hline
	\backslashbox{size}{ratio}
	&\makebox[6em]{small}&\makebox[6em]{big}\\\hline\hline
	small&room-like&corridor-like\\\hline
	big&hall-like&hall-like\\\hline
	
	\end{tabular}
	\caption{The classifier providing geometry evidence.}
	\label{TAB:classifier}
\end{table}

\begin{figure}
	\centering
	\includegraphics[width=0.7\columnwidth]{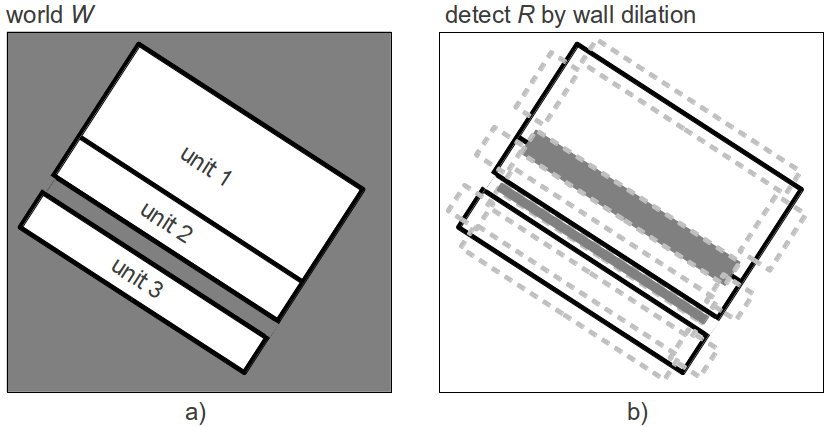}
    \caption{An example of relation detection. \textbf{a)} A semantic world $W$ containing three units (black=wall, white=free, gray=unknown). \textbf{b)} All four walls of each unit are dilated, with dashed rectangles in light-gray representing the dilated walls. The overlap of the dilated walls is shown in dark-gray which indicates the relation of ``adjacent". The overlap is detected using connected-components analysis \cite{chang04}. In this example, unit 1 and unit 3 are not adjacent; unit 2 and unit 3 are adjacent; unit 1 and unit 2 are adjacent.}
    \label{figure:detect-r}
\end{figure}

Relation evidence is detected based on image processing techniques: we first dilate all four walls of each unit, and then relation $r_{p,q}$ between the unit $u_p$ and $u_q$ is decided according to connected-components analysis \cite{chang04}. An example of relation detection is depicted in Fig.      \ref{figure:detect-r}, where $R$ is given by
\begin{equation}
    R=
    \left[\begin{array}{ccc}
    \neg\textrm{adj}&\textrm{adj}&\neg\textrm{adj}\\
    \textrm{adj}&\neg\textrm{adj}&\textrm{adj}\\
    \neg\textrm{adj}&\textrm{adj}&\neg\textrm{adj}\\
    \end{array}\right].
\end{equation}
Similar to relation detection, doors are detected as small open line segments which are located on the connecting walls of two neighbour space units. Details on door detection can be found in \cite{liu2013extracting}.

Given necessary evidence, we can make inference in MLNs and use the inference results to calculate the prior and likelihood. In this work, we have used the ProbCog Toolbox \cite{probcog} to perform MLN inference. Currently, we use hard evidences for knowledge processing, however, our system is also able to process soft evidences, as long as the evidences are provided in the soft form.

%


\subsection{Inference-based prior and likelihood design}
\subsubsection{Prior}
\indent  According to the model definition in equation (\ref{eq:abstract model}), the prior $p(W)$ is given by
\begin{eqnarray}
p(W)&=&p(U,T,R)\nonumber\\
&=&p(U|T,R) \cdot p(T,R).
\end{eqnarray}
Here, $p(U|T,R)$ can be seen as a factor expressing the dependency of the geometry parameters of the underlying units (see Fig. \ref{figure:system overview}-d) on the abstract terms in the MLNs. In our case, the geometry (size, position and orientation) of a unit is described by its four vertices. Furthermore, we define 
\begin{equation}
p(U|T,R):=\eta\prod_{p,q\in [1,2,\dots,n]}b(u_p,u_q), \label{equ:same_length}
\end{equation}
with 
\begin{equation}
b(u_p,u_q)=\left\{\begin{array}{lcc}
e^{-\frac{d^2}{2\sigma^2}}, p(\textit{SaLe}(u_p,u_q)|\text{evidence})>\textrm{threshold}\\\quad ~~~~~~\textrm{and}~~p\neq q,\\
1,~~ \text{otherwise}, 
\end{array}
\right.\label{eq:prior}
\end{equation}
where $n$ is the total number of units, and $d$ represents the length difference of the connecting walls of two adjacent units. $e^{-\frac{d^2}{2\sigma^2}}$ indicates a Gaussian function with mean at zero. $p(\textit{SaLe}(u_p,u_q)|\text{evidence})$ is one of the inferences that we can make in MLNs. $\eta$ is the normalization factor which ensures that $p(U|T,R)$ integrates to one. At the current stage, we assume that $p(T,R)$ follows a uniform distribution. However, it is possible to learn this distribution given proper training data.

\begin{figure}[!htb]
	\centering
	\includegraphics[width=.6\columnwidth]{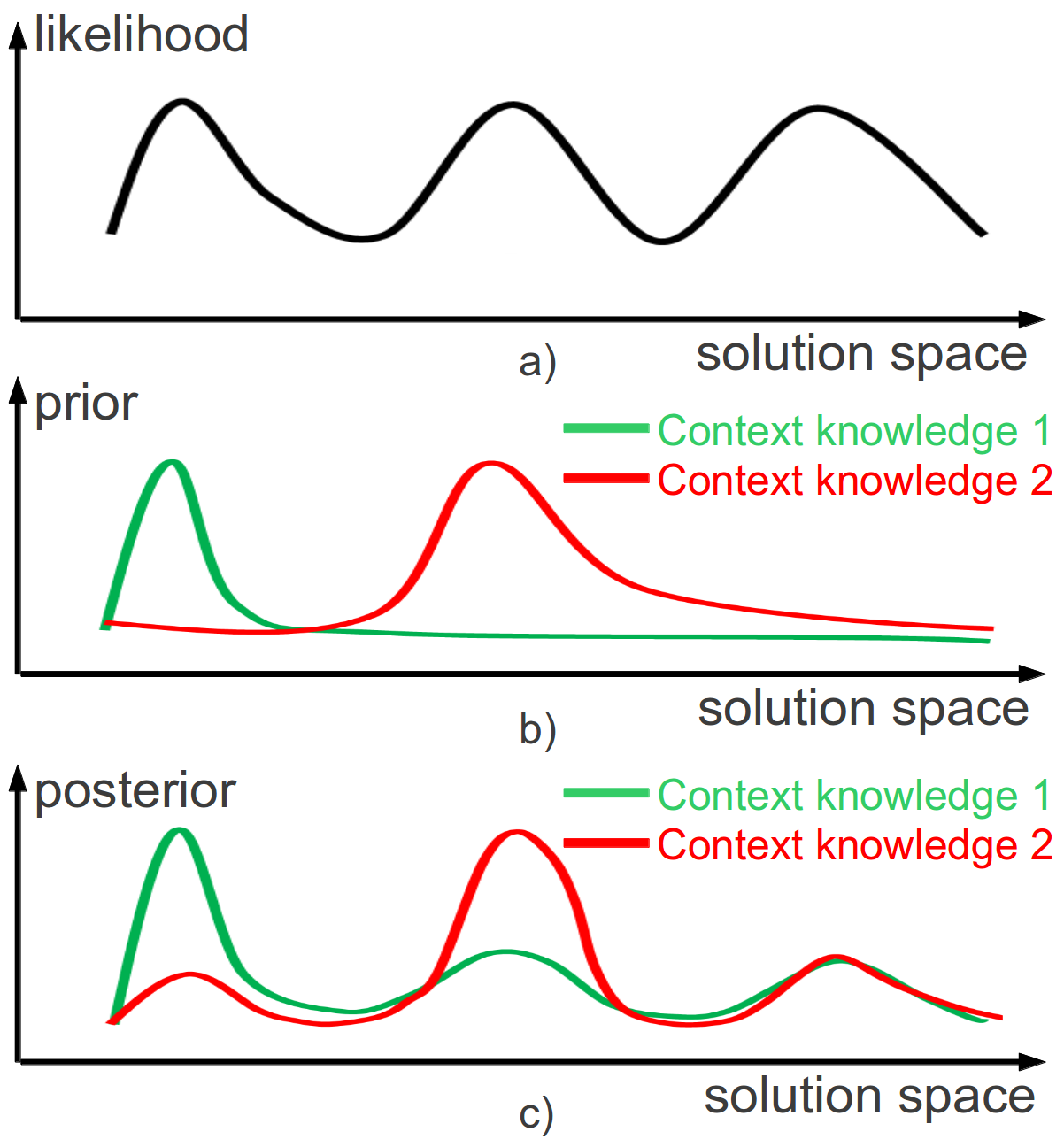}
    \caption{The general concept of inference-based prior design illustrated using a one-dimensional example. \textbf{a)} The likelihood for different settings of models, which contains three optima. \textbf{b)} The prior distribution represented by context knowledge defined in MLNs. Different context knowledge (set of rules) represents different prior distribution (green and red). If no context knowledge is used, it is the same as implementing the context knowledge that represents a uniform distribution which does not influence the posterior, i.e. posterior is only proportional to likelihood. \textbf{c)} Corresponding posterior distributions obtained using the two prior distributions shown in figure b. By setting prior distribution by inference, we shape the posterior so that the number of optima decreases, which means, the models complying with our context knowledge tend to have high posterior probability.
}
    \label{figure:posterior concept}
\end{figure}

\indent So far, the prior $p(W)$ is defined based on the inference results of MLNs, which enforces that the semantic worlds that comply with the context knowledge have high prior probability. Note that the worlds that contradict the context knowledge are not given a zero prior probability, instead, they become less probable. The general idea of inference-based prior design is explained in Fig. \ref{figure:posterior concept}, using a one-dimensional example. 

\subsubsection{Likelihood}
\indent Let $c(x,y)$ be the grid cell with the coordinate $(x,y)$ in the input map $M$, then we define the likelihood $p(M|W)$ as follows:
\begin{equation}
p(M|W)=\prod\limits_{c(x,y)\in M}  \alpha(c(x,y))\cdot p(c(x,y)|W).
\label{eq:likelihood}
\end{equation}
Here $\alpha(c(x,y))$ penalizes overlap between units and is given by
\begin{equation}
\alpha(c(x,y))=\psi^{\gamma(c(x,y))},
\end{equation}
with
\begin{eqnarray}
\gamma(c(x,y))&=&\left\{\begin{array}{lc}
\sigma(c(x,y))-1,\sigma(c(x,y))>1\\
0,\textrm{otherwise},\\
\end{array}
\right. 
\end{eqnarray}
where $\psi$ is a penalization factor with $\psi\in(0,1)$. $\sigma(c(x,y))$ indicates the number of units, to which $c(x,y)$ belongs.


\indent In equation (\ref{eq:likelihood}), the term $p(c(x,y)|W)$ is a semantic sensor model and evaluates the match between the world $W$ and input map $M$. Essentially, $p(c(x,y)|W)$ captures the quality of the original mapping algorithm producing the grid map which is used as input in our system. For calculating $p(c(x,y)|W)$, we discretize the cell state $M(x,y)$ of the input map into three classes ``occupied", ``unknown" and ``free", by thresholding its occupancy values. Our semantic world $W$ contains four types of cell states, which are
\begin{itemize}
\item ``wall": cells on the four edges of each unit.
\item ``object": cells that are located within a unit and are considered as non-free. These cells are detected using connected-components analysis \cite{chang04}.
\item ``free": cells that are located within a unit and do not belong to the class ``object".
\item ``unknown": cells that are located outside all units.

\end{itemize}
In this way, our semantic sensor model $p(c(x,y)|W)$ is realized as a ``3$\times$4" look-up table. 

\begin{figure}[!htb]
	\centering
	\includegraphics[width=.9\columnwidth]{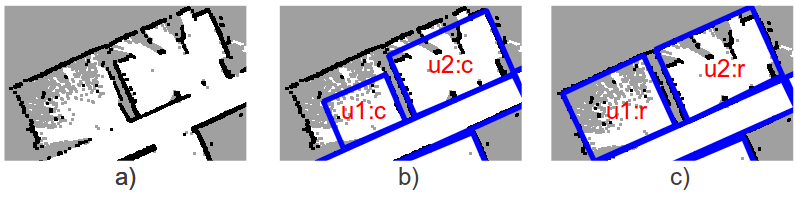}
    \caption{Effect of our semantic sensor model. \textbf{a)} The input map. \textbf{b)} The highest likelihood solution for unit 1 and unit 2 if they are of the type ``corridor". \textbf{c)} The highest likelihood solution for unit 1 and unit 2 if they are of the type ``room". Note that the type of a unit is decided based on the inference results in MLNs (Table \ref{TAB:formulas}).}
    \label{figure:semantic sensor model}
\end{figure}

In the real world, it is more likely that rooms and halls contain objects than corridors do, because the functionality of corridors is connecting other units, rather than placing objects. Thus, we propose to make the values of our semantic sensor model dependent on the type (decided based on the inference results in MLNs) of the underlying unit. The fundamental idea is to set the values of the semantic sensor model for the units of non-corridor types in such a way that it does not strongly penalize the mismatch between the input and the semantic world, and thus allows the existence of false positives (potential object cells). The effect of our semantic sensor model is depicted in Fig. \ref{figure:semantic sensor model}.

\section{Stochastic Generation of Semantic Worlds}
\begin{figure}[!htb]
	\centering
  	\includegraphics[width=.9\columnwidth]{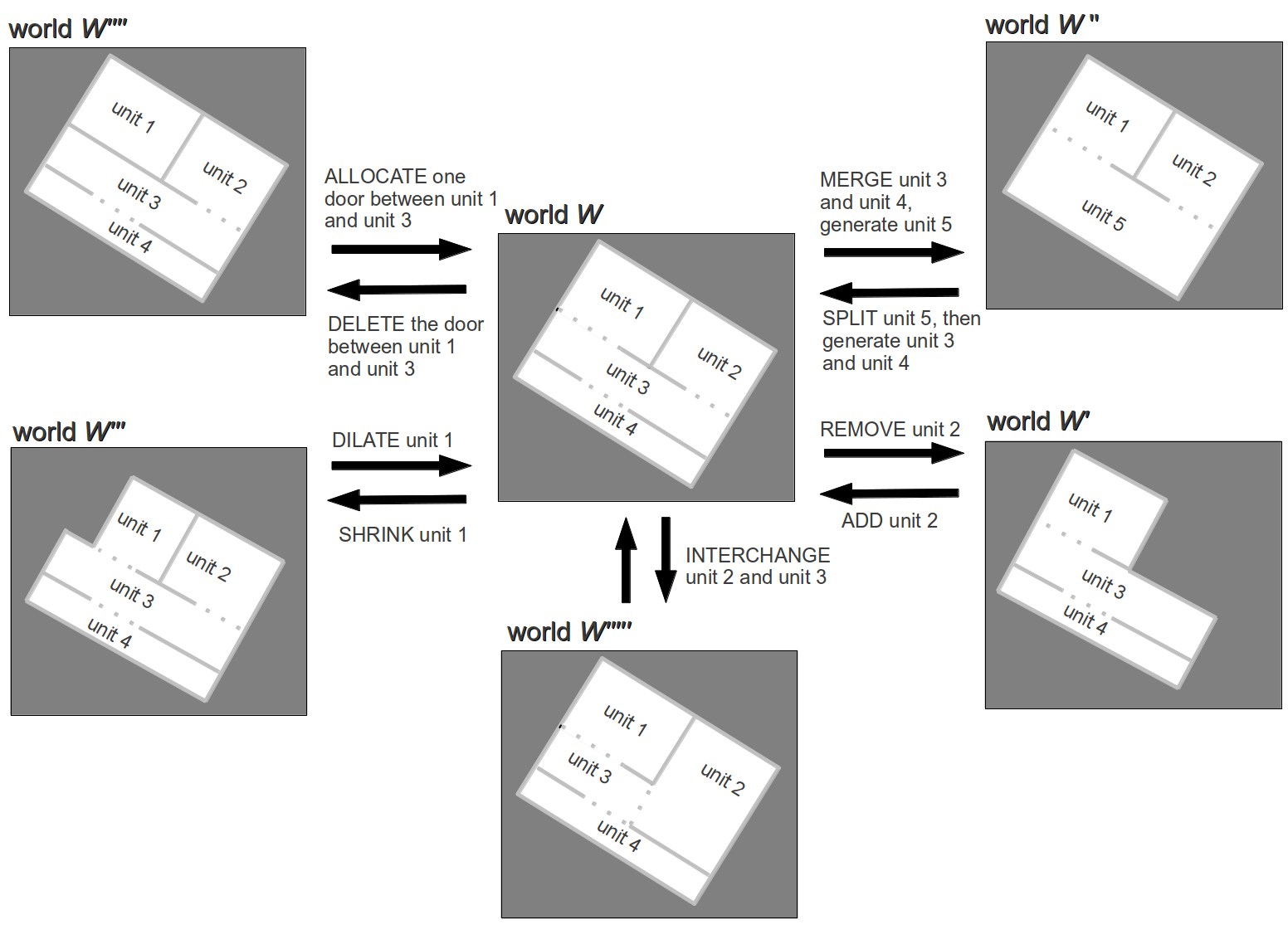}.
    \caption{Reversible MCMC kernels: ADD/REMOVE, SPLIT/MERGE, SHRINK/DILATE, ALLOCATE/DELETE and INTERCHANGE. The world $W$ can transit to $W^{'}$, $W^{''}$, $W^{'''}$, $W^{''''}$ and $W^{'''''}$ by applying the kernel REMOVE, MERGE, SHRINK, DELETE and INTERCHANGE, respectively. By contrast, the world $W^{'}$, $W^{''}$, $W^{'''}$, $W^{''''}$ and $W^{'''''}$ can also transit back to $W$ using corresponding reverse kernel.
}
   	\label{figure:mcmc-kernels-knowledge}
\end{figure}

\indent Given the posterior probability $p(W|M)$ defined in equation (\ref{eq:pos}), we aim to obtain the \textit{maximum a posteriori} solution $W^*$:
\begin{equation}
W^*=\arg\!\max_{\!\!\!\!\!\!\!\!\!\!\!_{W\in\Omega}} \, p(W|M),
\label{eq:argmax-w}
\end{equation}
where $\Omega$ indicates the solution space of semantic worlds. In order to find $W^*$, we use a data driven MCMC sampling technique \cite{tu2005image}. This technique constructs a Markov chain, and each state of this Markov chain represents a semantic world. By sequentially applying transition kernels to the current world (Fig. \ref{figure:mcmc-kernels-knowledge}) and accepting this transition by certain probability, this technique is able to efficiently draw samples from the corresponding posterior distribution. In addition to the four reversible kernel pairs that are defined in our previous work \cite{liu2013extracting}, we propose here a new reversible kernel ``INTERCHANGE" that changes the structure of two adjacent units at the same time, without changing the total size. These structural changes are proposals that allow our system to escape local optima. Fig. \ref{figure:mcmc-kernels-knowledge} shows an example of the reversible MCMC kernels. More details on the realization of the  used data driven MCMC process can be found in \cite{liu2013extracting}.

\section{Experiments and Discussion}
\label{exp}
\begin{figure*}[!htb]
	\centering
	\includegraphics[width=1.6\columnwidth]{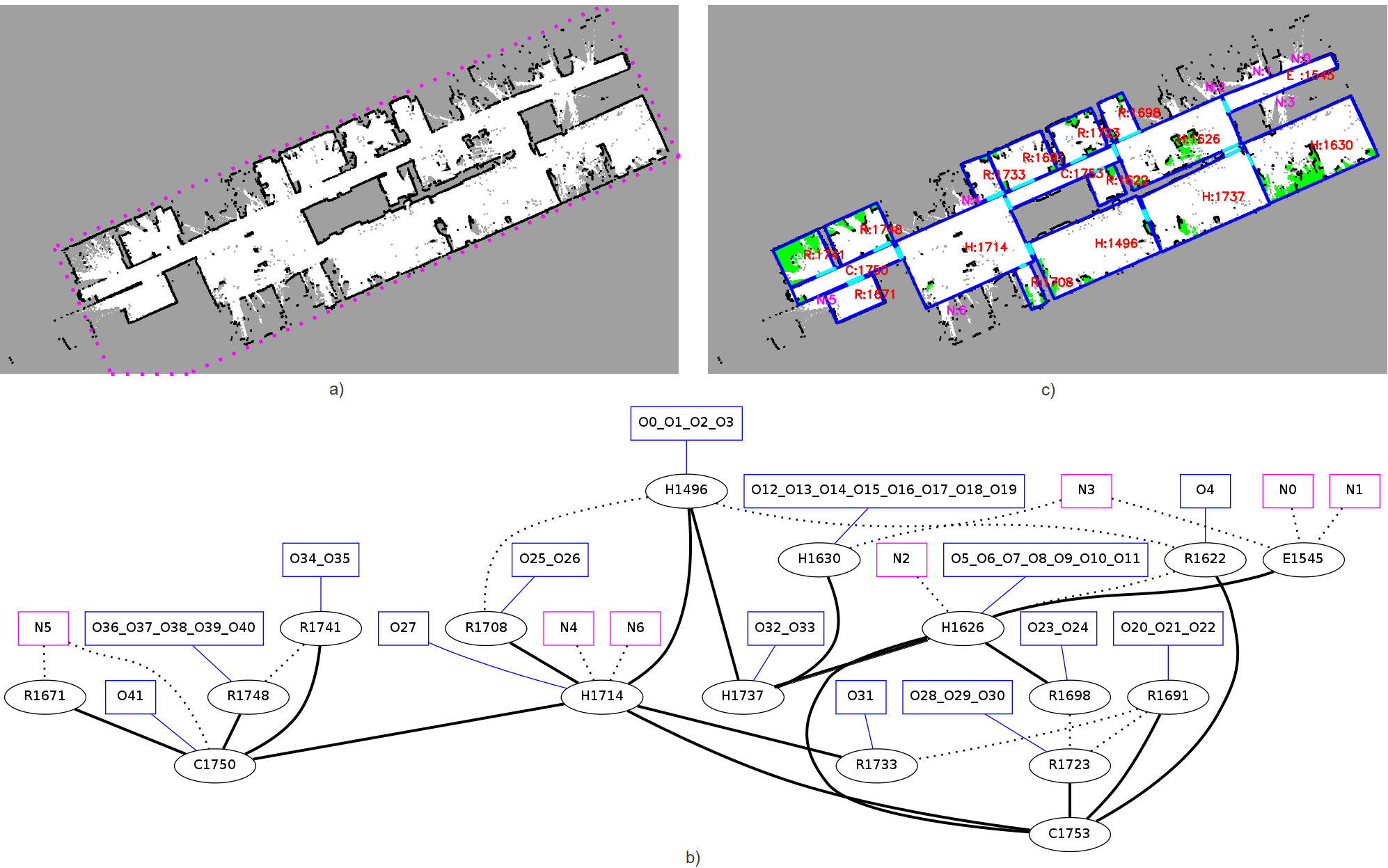}
    \caption{The performance of our semantic mapping system. \textbf{a)} The classified input map $C_M$ (black=occupied, gray=unknown, white=free). The manually defined region of interest (marked by magenta dashed lines) is used only for quantitative evaluation. \textbf{b)} The corresponding abstract model of $W^*$ (ellipse node=space unit with ID, blue rectangle node=detected objects with ID, magenta rectangle node=unexplored area with ID, black solid edge=connected by door, black dashed edge=adjacent without door). \textbf{c)} The most likely semantic world $W^*$ plotted onto the classified map $C_M$ (black=occupied, blue=wall, gray=unknown, white=free, cyan=door, green=object). The type and ID of each unit is shown at its center (R=room, H=hall, C=corridor, E=other). Unexplored area is detected using connected-components analysis \cite{chang04} and is marked by magenta ``N". These areas are too small to be recognized as space unit but are evidence for physically existing space.}
    \label{figure:performance}
\end{figure*}

\indent In this paper, we extend our previous work \cite{liu2013extracting} with inference using rule-based context knowledge, and the performance of our current system is shown in Fig. \ref{figure:performance}. As input, a big occupancy grid map $M$ (``ubremen-cartesium" dataset \cite{Radish}) of an entire floor of a building is used. This map is a big matrix (``1237$\times$672") containing occupancy values. We classify these values as $\{occupied,unknown,free\}$ so as to generate the classified input map $C_M$ (Fig. \ref{figure:performance}-a). Starting from a random initial guess, the semantic world $W$ is adapted to better match the input map $M$ by stochastically applying the kernels shown in Fig. \ref{figure:mcmc-kernels-knowledge}. After certain burn-in time, we get the most likely semantic world $W^*$ comprised of 17 units (each of which is represented by a rectangle) as shown in Fig. \ref{figure:performance}-c. Not only does $W^*$ accurately represent the geometry of the input map, but also $W^*$ is a parametric abstract model (Fig. \ref{figure:performance}-b) of the input map that provides valuable abstract information, such as adjacency, existence of objects and connectivity by doors. In addition, unexplored areas are also captured by our abstract model (marked by magenta ``N"). These areas are too small to be recognized as space units but are evidence for physically existing space.

\begin{figure*}[!htb]
	\centering
	\includegraphics[width=1.6\columnwidth]{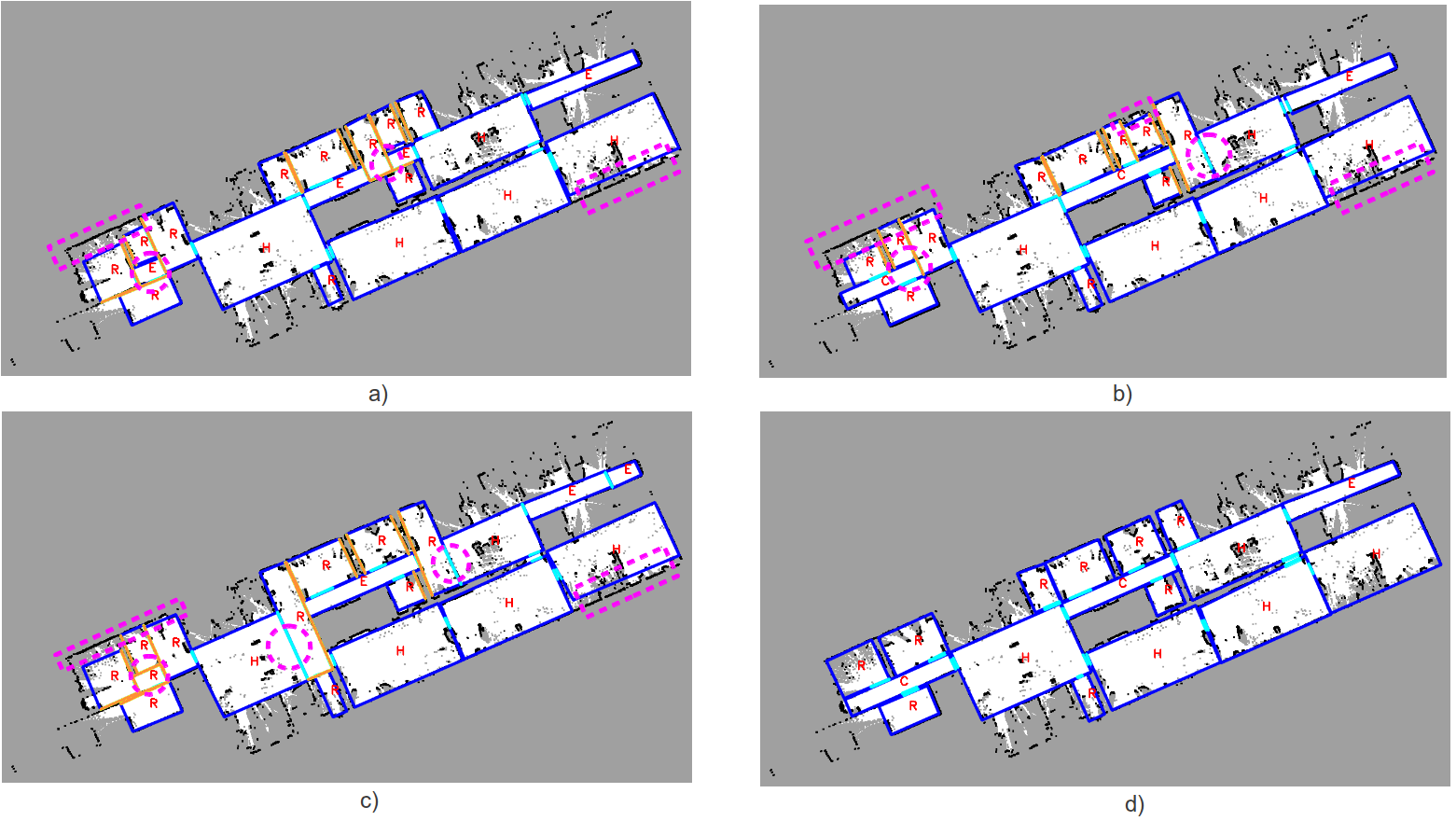}
    \caption{Comparison of performance between our previous work \cite{liu2013extracting} (figure a, b and c) and current system (figure d). Topological defects of the previous results are highlighted by magenta dashed circles, and poor local matches that are improved by semantic sensor model are highlighted by magenta dashed rectangles. Each pair of connecting walls that should have the same length are drawn in orange (figure a, b and c). Type of each unit is shown at its center.}
    \label{figure:comparison1}
\end{figure*}

\indent Compared with our previous work \cite{liu2013extracting}, our current system employs context knowledge in a systematic way, so that the input map is explained according to the underlying model structure. A performance comparison between our previous work and our current system is depicted in Fig.~\ref{figure:comparison1}. Three high likelihood samples obtained from our previous work are shown in Fig.~\ref{figure:comparison1}-a,b,c. They essentially represent the local maxima of the likelihood shown in  Fig.~\ref{figure:posterior concept}. Although all these three results provide good match to the input map (in terms of high likelihood), they have topological defects (highlighted by magenta circles), which contradict our knowledge (low prior). In this case, all pairs of connecting walls of adjacent rooms that should have the same length are drawn in orange in Fig. \ref{figure:comparison1}-a,b,c. The length difference of each pair of these connecting walls results in a penalization in prior probability (equation (\ref{eq:prior})). By applying our rule based context knowledge, local maxima of the likelihood with topological defects are suppressed so that they have a low posterior probability. In this way a semantic world that has high likelihood and high prior, i.e. high posterior (Fig.~\ref{figure:comparison1}-d), is easily obtained by stochastic sampling. In addition, various poor local matches (highlighted by magenta rectangles in Fig. \ref{figure:comparison1}-a,b,c) are corrected by the semantic sensor model used in our current system.

\begin{figure}[!htb]
	\centering
	\includegraphics[width=.8\columnwidth]{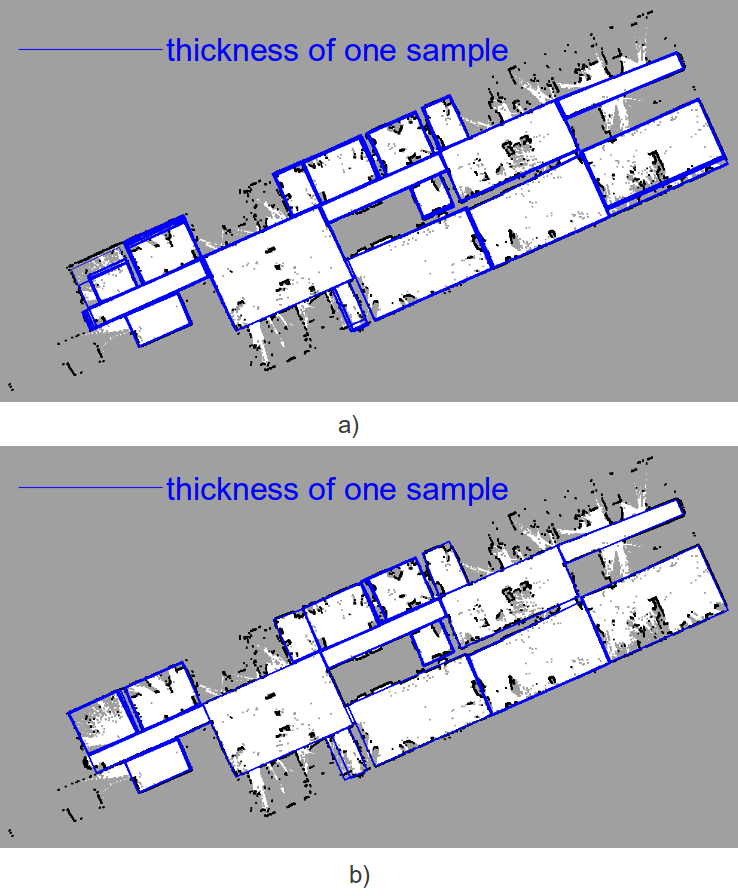}
    \caption{Posterior distribution built by 1000 samples. Starting from the world state shown in figure \ref{figure:performance}-c, we plot 1000 accepted samples obtained from our previous work (figure a) and those obtained from the current system (figure b) onto the classified map . It is obvious that the underlying Markov chain converges better because of the incorporated rule-based knowledge defined in MLNs. Here, we purposefully draw each sample using very thin line.}
    \label{figure:distribution}
\end{figure}

\indent Starting the Markov chain from the world state $W^*$ as shown in Fig. \ref{figure:performance}-c, we show the posterior distribution obtained from our previous work (Fig.~\ref{figure:distribution}-a) and that obtained from our current system (Fig. \ref{figure:distribution}-b) by plotting 1000 accepted samples together. Here we purposefully plot each sample (world) using very thin line. It is obvious that the underlying Markov chain obtained from our current system is more stable and converges better (smaller variance).

\begin{figure}[!htb]
	\centering
	\includegraphics[width=.8\columnwidth]{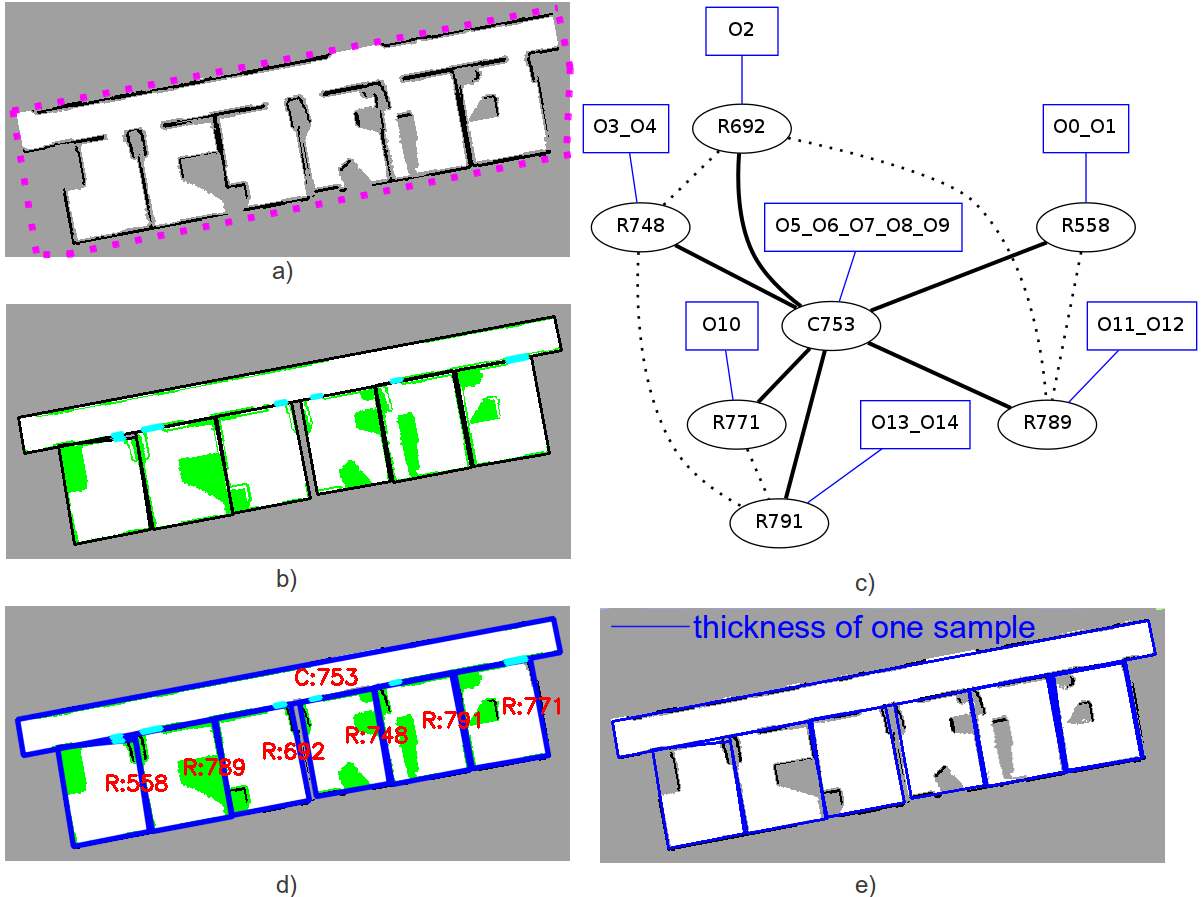}
    \caption{The performance of our current system for another data set. \textbf{a)} The classified input map. The manually defined region of interest (marked by magenta dashed lines) is used only for quantitative evaluation. \textbf{b)} The most likely semantic world. \textbf{c)} The corresponding abstract model. Here, no unexplored area is detected. \textbf{d)} A direct comparison between the input map and the corresponding semantic world. \textbf{e)} Posterior distribution built by 1000 samples. It is obvious that the underlying Markov chain achieves stable convergence using our current system.}
    \label{figure:intel}
\end{figure}

\indent Fig. \ref{figure:intel} shows the performance of our current system on another data set. As can be seen, our system accurately represents the geometry of the environments captured in the input maps and provides a semantic world that explains the perceived environments with the correct topology. We evaluate our system quantitatively using the measure ``cell prediction rate" (CPR), which denotes the percentage of the correctly explained cells in the manually defined region of interest (see Fig. \ref{figure:performance}-a and \ref{figure:intel}-a). The CPR of Fig. \ref{figure:performance}-c is 86.8\%, and that of Fig. \ref{figure:intel}-d is 91.4\%.

By modelling context knowledge in MLNs, we can assign semantic information, e.g. the type, to the data. This allows us to use a semantically informed sensor model to better explain the observations. Moreover, the used context knowledge, given by the rules, shape our prior so that unlikely configurations can be ruled out, as shown in Fig. \ref{figure:comparison1}.

The computational cost is strongly dependent on the size of the input map and consists of two parts, which are MCMC operations and knowledge processing in MLNs. With a single-threaded implementation on an Intel i7 CPU, the speed of MCMC operations is around 30 iterations per second for the map shown in Fig. \ref{figure:performance}. The speed of knowledge processing in MLNs depends on one hand on the tool (the software implementation of MLNs) that one uses. On the other hand, it depends on the number of optimization iterations set in the tool. In our case, we could get a satisfactory result in 5-8 seconds using the tool in \cite{probcog} per processing. To analyze a grid map, we first start our system without activating knowledge processing, only after enough context is available (e.g. coverage of the input map greater than 80\%), knowledge processing is enabled to help better explain the input map. In this way, we could obtain a good result within a reasonable processing time, which is, 20 minutes for the map shown in Fig. \ref{figure:performance}.

\section{Conclusion}
\label{sum}
\indent In this paper, we extended our previous work \cite{liu2013extracting} with inference using rule-based context knowledge. Our current system demonstrates an advanced stochastic sampling process supervised by the context knowledge which is defined as descriptive rules in Markov Logic Networks. As output, our system returns a parametric abstract model of the perceived environment that not only accurately represents the environment geometry, but also provides valuable abstract information, which serves as a basis for higher level reasoning processes. By constructing the prior distribution 
of the semantic maps using inference results, high likelihood results with topological defects (contradiction with context knowledge) are suppressed.  Furthermore, by applying a semantically annotated 
sensor model for likelihood calculation, we explicitly use the extracted semantic information to improve the performance of our system. 

\section*{Acknowledgements}

This work is accomplished with the support of the Technische Universit\"at M\"unchen - Institute for Advanced Study, funded by the German Excellence Initiative.

The work of one of the authors (Georg von Wichert) was partially made possible by funding from the ARTEMIS Joint Undertaking as part of the project R3-COP and from the German Federal Ministry of Education and Research (BMBF) under grant no. 01IS10004E.

\bibliographystyle{plain}
\bibliography{knowledge_iros}

\begin{thebibliography}{10}

\bibitem{an2012}
S.Y. An, L.K. Lee, and S.Y. Oh.
\newblock Fast incremental 3d plane extraction from a collection of 2d line
  segments for 3d mapping.
\newblock In {\em IEEE/RSJ International Conference on Intelligent Robots and
  Systems}. IEEE, 2012.

\bibitem{chang04}
Fu~Chang, Chun jen Chen, and Chi jen Lu.
\newblock A linear-time component-labeling algorithm using contour tracing
  technique.
\newblock {\em Computer Vision and Image Understanding}, 93:206--220, 2004.

\bibitem{goerke2009building}
N.~Goerke and S.~Braun.
\newblock Building semantic annotated maps by mobile robots.
\newblock In {\em Proceedings of the Conference Towards Autonomous Robotic
  Systems}, 2009.

\bibitem{grisetti2007improved}
G.~Grisetti, C.~Stachniss, and W.~Burgard.
\newblock Improved techniques for grid mapping with rao-blackwellized particle
  filters.
\newblock {\em IEEE Transactions on Robotics}, 23(1):34--46, 2007.

\bibitem{Radish}
A.~Howard and N.~Roy.
\newblock The robotics data set repository (radish), 2003.

\bibitem{probcog}
D.~Jain.
\newblock Probcog toolbox, http://ias.cs.tum.edu/software/probcog, 2011.

\bibitem{jebari2011multi}
I.~Jebari, S.~Bazeille, E.~Battesti, H.~Tekaya, M.~Klein, A.~Tapus, D.~Filliat,
  C.~Meyer, R.~Benosman, E.~Cizeron, et~al.
\newblock Multi-sensor semantic mapping and exploration of indoor environments.
\newblock In {\em IEEE Conference on Technologies for Practical Robot
  Applications (TePRA)}, pages 151--156. IEEE, 2011.

\bibitem{krishnan2010visual}
A.K. Krishnan and K.M. Krishna.
\newblock A visual exploration algorithm using semantic cues that constructs
  image based hybrid maps.
\newblock In {\em IEEE/RSJ International Conference on Intelligent Robots and
  Systems}, pages 1316--1321. IEEE, 2010.

\bibitem{liu2013extracting}
Z.~Liu and G.~von Wichert.
\newblock Extracting semantic indoor maps from occupancy grids.
\newblock {\em Robotics and Autonomous Systems}, 2013.

\bibitem{masonobject}
J.~Mason and B.~Marthi.
\newblock An object-based semantic world model for long-term change detection
  and semantic querying.
\newblock In {\em IEEE/RSJ International Conference on Intelligent Robots and
  Systems}. IEEE, 2012.

\bibitem{Nieto-Granda2010}
C.~Nieto-Granda, J.G. Rogers, A.J.B. Trevor, and H.I. Christensen.
\newblock Semantic map partitioning in indoor environments using regional
  analysis.
\newblock In {\em IEEE/RSJ International Conference on Intelligent Robots and
  Systems}, pages 1451--1456. IEEE, 2010.

\bibitem{nuechter2008towards}
A.~N{\"u}chter and J.~Hertzberg.
\newblock Towards semantic maps for mobile robots.
\newblock {\em Robotics and Autonomous Systems}, 56(11):915--926, 2008.

\bibitem{pangercicsemantic}
D.~Pangercic, B.~Pitzer, M.~Tenorth, and M.~Beetz.
\newblock Semantic object maps for robotic housework-representation,
  acquisition and use.
\newblock In {\em IEEE/RSJ International Conference on Intelligent Robots and
  Systems}. IEEE, 2012.

\bibitem{park2011hybrid}
J.T. Park and J.B. Song.
\newblock Hybrid semantic mapping using door information.
\newblock In {\em 8th International Conference on Ubiquitous Robots and Ambient
  Intelligence (URAI)}, pages 128--130. IEEE, 2011.

\bibitem{persson2007probabilistic}
M.~Persson, T.~Duckett, C.~Valgren, and A.~Lilienthal.
\newblock Probabilistic semantic mapping with a virtual sensor for
  building/nature detection.
\newblock In {\em International Symposium on Computational Intelligence in
  Robotics and Automation}, pages 236--242. IEEE, 2007.

\bibitem{pronobis2012large}
A.~Pronobis and P.~Jensfelt.
\newblock Large-scale semantic mapping and reasoning with heterogeneous
  modalities.
\newblock In {\em IEEE International Conference on Robotics and Automation},
  pages 3515--3522. IEEE, 2012.

\bibitem{ranganathan2007semantic}
A.~Ranganathan and F.~Dellaert.
\newblock Semantic modeling of places using objects.
\newblock In {\em Robotics: Science and Systems}, 2007.

\bibitem{richardson2006markov}
M.~Richardson and P.~Domingos.
\newblock Markov logic networks.
\newblock {\em Machine learning}, 62(1):107--136, 2006.

\bibitem{rusu2009model}
R.B. Rusu, Z.C. Marton, N.~Blodow, A.~Holzbach, and M.~Beetz.
\newblock Model-based and learned semantic object labeling in 3d point cloud
  maps of kitchen environments.
\newblock In {\em IEEE/RSJ International Conference on Intelligent Robots and
  Systems}, pages 3601--3608. IEEE, 2009.

\bibitem{shim20113d}
I.~Shim, Y.~Choe, and M.J. Chung.
\newblock 3d mapping in urban environment using geometric featured voxel.
\newblock In {\em International Conference on Ubiquitous Robots and Ambient
  Intelligence}, pages 804--805. IEEE, 2011.

\bibitem{sjoo2012semantic}
K.~Sjoo.
\newblock Semantic map segmentation using function-based energy maximization.
\newblock In {\em IEEE International Conference on Robotics and Automation},
  pages 4066--4073. IEEE, 2012.

\bibitem{tu2005image}
Z.~Tu, X.~Chen, A.L. Yuille, and S.C. Zhu.
\newblock Image parsing: Unifying segmentation, detection, and recognition.
\newblock {\em International Journal of Computer Vision}, 63(2):113--140, 2005.

\bibitem{wolf2008semantic}
D.F. Wolf and G.S. Sukhatme.
\newblock Semantic mapping using mobile robots.
\newblock {\em IEEE Transactions on Robotics}, 24(2):245--258, 2008.

\end{thebibliography}



\end{document}